\documentclass[sigconf, nonacm]{acmart}

\setcopyright{none}
\acmConference{}{}{}
\acmYear{}
\copyrightyear{}
\acmDOI{}
\acmISBN{}

\usepackage{balance} %

\usepackage{graphicx}      %
\usepackage{subcaption}    %
\usepackage{tikz}
\usetikzlibrary{decorations.pathmorphing,automata,positioning,arrows.meta}

\usepackage{amsmath,booktabs,colortbl}
\usepackage{tabularx}

\usepackage{pgfplots}
\usepgfplotslibrary{groupplots}
\pgfplotsset{compat=1.17}

\usepackage{enumitem,float,algorithm,algpseudocode}
\usepackage{cleveref}

\newcommand{\pcIcon}[1]{\tikz[baseline=-0.38em,x=1.05em,y=1.05em]\fill[black!80] #1;}
\newsavebox{\pcKnightBox}\newsavebox{\pcKingBox}\newsavebox{\pcRookBox}
\sbox{\pcKnightBox}{\pcIcon{(-0.086,0.338)..controls(-0.069,0.348)and(-0.061,0.362)..(-0.061,0.38)..controls(-0.061,0.391)and(-0.064,0.403)..(-0.071,0.415)..controls(-0.078,0.428)and(-0.081,0.439)..(-0.081,0.449)..controls(-0.081,0.47)and(-0.068,0.487)..(-0.042,0.5)..controls(-0.004,0.464)and(0.016,0.436)..(0.02,0.416)..controls(0.023,0.396)and(0.025,0.383)..(0.025,0.378)..controls(0.025,0.357)and(0.017,0.344)..(0.002,0.338)--(0.005,0.338)..controls(0.005,0.338)and(0.004,0.338)..(0.003,0.338)..controls(0.242,0.186)and(0.372,-0.093)..(0.392,-0.5)--(-0.257,-0.5)..controls(-0.262,-0.475)and(-0.265,-0.45)..(-0.265,-0.424)..controls(-0.265,-0.352)and(-0.22,-0.285)..(-0.13,-0.222)..controls(-0.04,-0.159)and(0.005,-0.102)..(0.005,-0.052)..controls(0.005,-0.049)and(0.005,-0.045)..(0.005,-0.042)..controls(-0.025,-0.117)and(-0.066,-0.154)..(-0.118,-0.154)..controls(-0.118,-0.154)and(-0.12,-0.154)..(-0.124,-0.154)..controls(-0.164,-0.219)and(-0.207,-0.251)..(-0.253,-0.251)..controls(-0.255,-0.251)and(-0.257,-0.251)..(-0.258,-0.251)--(-0.226,-0.182)--(-0.297,-0.235)..controls(-0.348,-0.223)and(-0.38,-0.198)..(-0.392,-0.161)..controls(-0.352,-0.059)and(-0.333,0.034)..(-0.333,0.119)..controls(-0.333,0.131)and(-0.333,0.143)..(-0.334,0.154)..controls(-0.313,0.203)and(-0.28,0.25)..(-0.236,0.297)..controls(-0.235,0.301)and(-0.234,0.305)..(-0.234,0.309)..controls(-0.234,0.322)and(-0.239,0.331)..(-0.249,0.337)..controls(-0.259,0.343)and(-0.264,0.353)..(-0.264,0.369)..controls(-0.264,0.385)and(-0.26,0.408)..(-0.251,0.438)..controls(-0.204,0.426)and(-0.169,0.393)..(-0.144,0.338)--cycle (-0.257,0.168)--(-0.257,0.157)..controls(-0.256,0.144)and(-0.253,0.137)..(-0.249,0.137)--(-0.235,0.137)..controls(-0.205,0.137)and(-0.189,0.148)..(-0.189,0.17)..controls(-0.189,0.185)and(-0.18,0.203)..(-0.16,0.225)--(-0.164,0.225)..controls(-0.207,0.225)and(-0.237,0.21)..(-0.253,0.182)--cycle (-0.33,-0.133)--(-0.33,-0.149)..controls(-0.33,-0.162)and(-0.324,-0.168)..(-0.311,-0.168)--(-0.305,-0.168)..controls(-0.273,-0.162)and(-0.257,-0.147)..(-0.257,-0.123)--(-0.261,-0.1)..controls(-0.289,-0.1)and(-0.312,-0.111)..(-0.33,-0.133)--cycle (0.003,0.285)--(0.037,0.225)..controls(0.205,0.105)and(0.289,-0.125)..(0.292,-0.465)--(0.344,-0.465)..controls(0.324,-0.086)and(0.21,0.164)..(0.003,0.285)--cycle}}
\sbox{\pcKingBox}{\pcIcon{(-0.047,0.146)..controls(-0.046,0.148)and(-0.046,0.149)..(-0.045,0.15)..controls(-0.06,0.17)and(-0.08,0.185)..(-0.102,0.198)..controls(-0.135,0.216)and(-0.169,0.225)..(-0.204,0.225)..controls(-0.239,0.225)and(-0.272,0.216)..(-0.305,0.198)..controls(-0.338,0.18)and(-0.364,0.155)..(-0.382,0.121)..controls(-0.401,0.088)and(-0.41,0.054)..(-0.41,0.018)..controls(-0.41,-0.018)and(-0.401,-0.052)..(-0.383,-0.085)..controls(-0.365,-0.117)and(-0.339,-0.143)..(-0.306,-0.16)..controls(-0.296,-0.167)and(-0.286,-0.172)..(-0.276,-0.175)--(-0.047,-0.175)--cycle (0.047,-0.175)--(0.276,-0.175)..controls(0.287,-0.172)and(0.297,-0.167)..(0.307,-0.16)..controls(0.34,-0.143)and(0.365,-0.117)..(0.383,-0.085)..controls(0.401,-0.052)and(0.41,-0.018)..(0.41,0.018)..controls(0.41,0.054)and(0.401,0.088)..(0.383,0.121)..controls(0.364,0.155)and(0.339,0.18)..(0.306,0.198)..controls(0.273,0.216)and(0.239,0.225)..(0.204,0.225)..controls(0.17,0.225)and(0.136,0.216)..(0.103,0.198)..controls(0.08,0.185)and(0.061,0.17)..(0.045,0.15)..controls(0.046,0.149)and(0.047,0.148)..(0.047,0.146)--cycle (-0.253,-0.244)--(-0.253,-0.447)--(0.253,-0.447)--(0.253,-0.244)--cycle (-0.307,-0.5)--(-0.307,-0.244)..controls(-0.326,-0.238)and(-0.344,-0.23)..(-0.361,-0.221)..controls(-0.392,-0.197)and(-0.419,-0.163)..(-0.443,-0.119)..controls(-0.467,-0.075)and(-0.479,-0.03)..(-0.479,0.018)..controls(-0.479,0.066)and(-0.467,0.112)..(-0.443,0.156)..controls(-0.418,0.2)and(-0.384,0.234)..(-0.34,0.257)..controls(-0.295,0.281)and(-0.25,0.293)..(-0.204,0.293)..controls(-0.168,0.293)and(-0.133,0.286)..(-0.1,0.272)--(-0.1,0.273)..controls(-0.1,0.289)and(-0.096,0.304)..(-0.088,0.319)..controls(-0.08,0.333)and(-0.068,0.344)..(-0.054,0.352)..controls(-0.048,0.355)and(-0.042,0.357)..(-0.035,0.359)--(-0.035,0.395)--(-0.107,0.395)--(-0.107,0.442)--(-0.035,0.442)--(-0.035,0.5)--(0.035,0.5)--(0.035,0.442)--(0.108,0.442)--(0.108,0.395)--(0.035,0.395)--(0.035,0.359)..controls(0.043,0.357)and(0.049,0.355)..(0.055,0.352)..controls(0.069,0.344)and(0.08,0.333)..(0.088,0.319)..controls(0.096,0.304)and(0.1,0.289)..(0.1,0.273)--(0.1,0.272)..controls(0.134,0.286)and(0.169,0.293)..(0.204,0.293)..controls(0.25,0.293)and(0.295,0.281)..(0.34,0.257)..controls(0.384,0.234)and(0.418,0.2)..(0.443,0.156)..controls(0.467,0.112)and(0.479,0.066)..(0.479,0.018)..controls(0.479,-0.03)and(0.467,-0.075)..(0.443,-0.119)..controls(0.419,-0.163)and(0.392,-0.197)..(0.362,-0.221)..controls(0.344,-0.23)and(0.326,-0.238)..(0.308,-0.244)--(0.308,-0.5)--cycle (-0.062,0.268)..controls(-0.062,0.257)and(-0.059,0.247)..(-0.054,0.237)..controls(-0.048,0.227)and(-0.04,0.22)..(-0.03,0.214)..controls(-0.02,0.209)and(-0.01,0.206)..(0,0.206)..controls(0.011,0.206)and(0.021,0.209)..(0.031,0.214)..controls(0.041,0.22)and(0.049,0.227)..(0.054,0.237)..controls(0.059,0.247)and(0.062,0.257)..(0.062,0.268)..controls(0.062,0.279)and(0.059,0.289)..(0.054,0.299)..controls(0.049,0.309)and(0.041,0.317)..(0.031,0.322)..controls(0.021,0.327)and(0.011,0.33)..(0,0.33)..controls(-0.01,0.33)and(-0.02,0.327)..(-0.03,0.322)..controls(-0.04,0.317)and(-0.048,0.309)..(-0.054,0.299)..controls(-0.059,0.289)and(-0.062,0.279)..(-0.062,0.268)--cycle (0.219,-0.279)--(0.219,-0.412)--(-0.218,-0.412)--(-0.218,-0.279)--cycle (-0.08,-0.141)--(-0.267,-0.141)..controls(-0.275,-0.138)and(-0.283,-0.134)..(-0.291,-0.128)..controls(-0.318,-0.114)and(-0.339,-0.093)..(-0.354,-0.066)..controls(-0.368,-0.039)and(-0.376,-0.01)..(-0.376,0.019)..controls(-0.376,0.049)and(-0.368,0.077)..(-0.353,0.105)..controls(-0.338,0.132)and(-0.317,0.153)..(-0.29,0.168)..controls(-0.264,0.183)and(-0.236,0.191)..(-0.208,0.191)..controls(-0.18,0.191)and(-0.153,0.183)..(-0.126,0.168)..controls(-0.107,0.158)and(-0.092,0.145)..(-0.079,0.129)..controls(-0.08,0.128)and(-0.08,0.127)..(-0.08,0.126)--cycle (0.081,0.126)..controls(0.081,0.127)and(0.08,0.128)..(0.08,0.129)..controls(0.092,0.145)and(0.108,0.158)..(0.126,0.168)..controls(0.154,0.183)and(0.181,0.191)..(0.208,0.191)..controls(0.237,0.191)and(0.265,0.183)..(0.291,0.168)..controls(0.318,0.153)and(0.338,0.132)..(0.354,0.105)..controls(0.369,0.077)and(0.376,0.049)..(0.376,0.019)..controls(0.376,-0.01)and(0.369,-0.039)..(0.354,-0.066)..controls(0.34,-0.093)and(0.319,-0.114)..(0.291,-0.128)..controls(0.283,-0.134)and(0.275,-0.138)..(0.267,-0.141)--(0.081,-0.141)--cycle}}
\sbox{\pcRookBox}{\pcIcon{(0.214,0.274)--(0.214,0.337)--(-0.214,0.337)--(-0.214,0.274)--cycle (0.264,-0.379)--(0.264,-0.305)--(-0.263,-0.305)--(-0.263,-0.379)--cycle (-0.095,0.5)--(0.091,0.5)--(0.091,0.396)--(0.205,0.396)--(0.205,0.5)--(0.385,0.5)--(0.385,0.392)--(0.308,0.392)--(0.308,0.295)--(0.247,0.195)--(0.247,-0.196)--(0.308,-0.295)--(0.308,-0.391)--(0.385,-0.391)--(0.385,-0.5)--(-0.385,-0.5)--(-0.385,-0.391)--(-0.308,-0.391)--(-0.308,-0.295)--(-0.246,-0.194)--(-0.246,-0.194)--(-0.246,0.195)--(-0.246,0.195)--(-0.308,0.295)--(-0.308,0.392)--(-0.385,0.392)--(-0.385,0.5)--(-0.208,0.5)--(-0.208,0.396)--(-0.095,0.396)--cycle (0.349,-0.463)--(0.349,-0.426)--(-0.348,-0.426)--(-0.348,-0.463)--cycle}}
\DeclareRobustCommand{\pcKnight}{\usebox{\pcKnightBox}}
\DeclareRobustCommand{\pcKing}{\usebox{\pcKingBox}}
\DeclareRobustCommand{\pcRook}{\usebox{\pcRookBox}}

\title[Assume/Guarantee Contracts for MARL]{
Assume/Guarantee Contracts for the Compositional
Multi-Agent Reinforcement Learning of Heterogeneous Teams
}

\author{Dominic Klukas}
\affiliation{
  \institution{University of British Columbia}
  \city{Vancouver}
  \country{Canada}}
\email{klukasdh@student.ubc.ca}

\author{Cyrus Neary}
\affiliation{
  \institution{University of British Columbia}
  \city{Vancouver}
  \country{Canada}}
\email{cyrus.neary@ubc.ca}

\begin{abstract}

Multi-agent reinforcement learning (MARL) problems for heterogeneous teams require agents to collaborate to complete team tasks while making the best use of diverse agent capabilities. 
We introduce assume/guarantee contracts for heterogeneous multi-agent systems (AGC-MAS) and Iterative Heterogeneous Assignment for MARL (IHA-MARL).
AGC-MAS contracts are compositional objects that encode the assumptions provided to, and the guarantees required of, each subtask contributing to the overall team task. 
We formulate contracts as tuples that specify a subtask, an automaton, timeout and success-probability requirements, and dispatch behavior. 
The automaton tracks an agent's high-level progress through the subtask associated with its contract and provides a natural mechanism for synchronization across contracts. 
Our abstraction enables robust interaction across heterogeneous low-level subtask implementations, allows heterogeneous agents to substitute for one another in satisfying contracts, and supports reuse of previously satisfied contracts in new team contexts. 
Having defined our theoretical framework, we develop a decomposition algorithm that breaks down a team task into a set of contracts. 
We then propose an iterative algorithm for task assignment and MARL that alternates between assigning contracts representing subtasks to individual agents, training the assigned agents to satisfy their contracts, and assessing their resulting capabilities to inform the next round of task assignments until the team's success probability meets the desired threshold. 
Our experimental results demonstrate the versatility of contracts in enabling agent substitution and contract reuse. 
Across three benchmarks with five seeds each, IHA-MARL achieves observed
$100\%$ final evaluation success. In CPU timing pilots, comparable
baselines required approximately $7$--$11\times$ more training-loop time.
\end{abstract}


\newcommand{\BibTeX}{\rm B\kern-.05em{\sc i\kern-.025em b}\kern-.08em\TeX}

\begin{document}
\pagestyle{plain}

\maketitle

\section{Introduction}

By distributing themselves geographically, delegating skills to different specialized agents, and working in parallel at scale, cooperative multi-agent systems (MAS) are increasingly necessary to many diverse applications, from autonomous manufacturing and field robotics, to software-based tasks such as cooperating teams of LLM agents \cite{MAS_UTILITY, LLM_MAS}.
In particular, heterogeneous multi-agent systems, in which each cooperating agent may have different capabilities, are necessary to many applications in which individual agents must specialize to perform specific tasks. 
For instance, aerial drones excel in scouting and surveying tasks, but are often less suitable to transportation tasks, whereas ground robots can carry heavy loads much more efficiently.
Because heterogeneous agents may contribute to a team task in fundamentally different ways, effective coordination requires algorithms that reason explicitly about both which agents are suited to particular subtasks and how their distinct contributions should be integrated.

Nevertheless, multi-agent teams face multiple unique challenges.
First, monolithic approaches to multi-agent decision-making and learning suffer from the curse of dimensionality \cite{MARL_SURVEY}.
Moving toward distributed approaches, however, raises the problem of automatically decomposing a complex team task into subtasks representing the contributions of individual agents \cite{auto_decompose_RM}.
In heterogeneous multi-agent teams, task decomposition and assignment must also account for variation in agent performance across subtasks.
When these capabilities are not known a priori, effective assignment may first require costly exploration and training to discover which agents are suited to which subtasks.
Distributed multi-agent learning must also coordinate interacting agents under limited communication and decentralized execution, often while contending with non-stationarity induced by simultaneously adapting policies.

\begin{figure*}[t]
  \centering
  \usetikzlibrary{arrows.meta,positioning,calc,automata,decorations.pathmorphing}%
\definecolor{ovSpec}{HTML}{534AB7}%
\definecolor{ovSpecFill}{HTML}{EEEDFE}%
\definecolor{ovPanel}{HTML}{888780}%
\definecolor{ovImpl}{HTML}{F1EFE8}%
\definecolor{ovOk}{HTML}{3B6D11}%
\definecolor{ovBad}{HTML}{A32D2D}%
\ifdefined\ovMapBox\else\newsavebox{\ovMapBox}\newsavebox{\ovAutBox}\fi%
\sbox{\ovMapBox}{\resizebox{3.1cm}{!}{\input{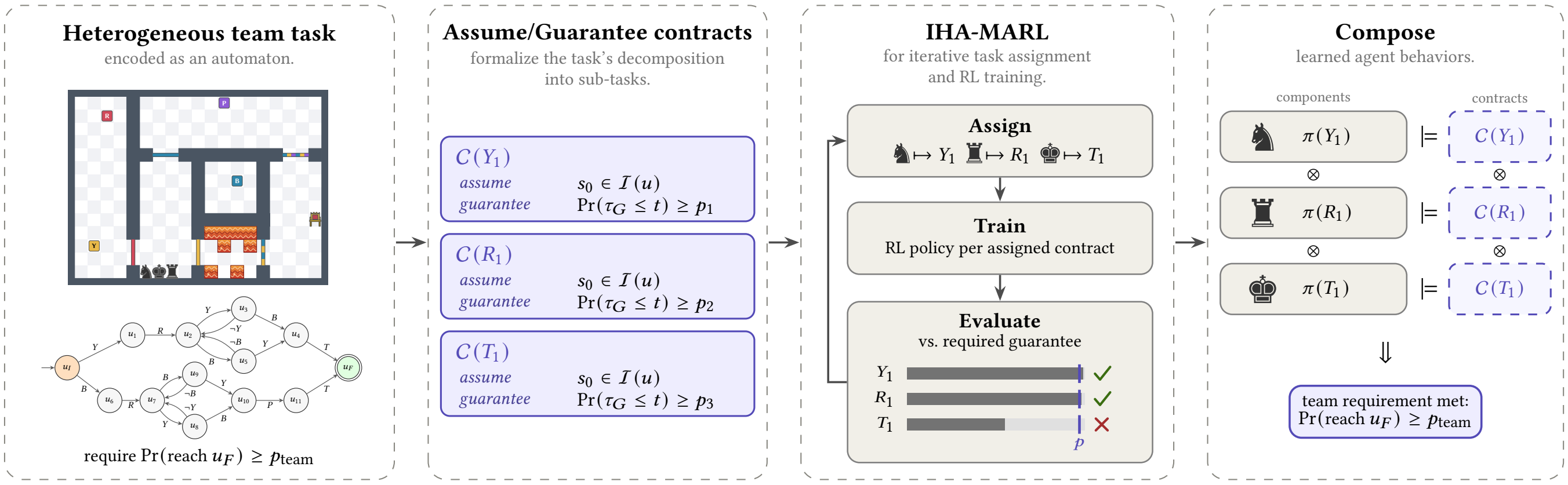}}}%
\sbox{\ovAutBox}{\resizebox{3.9cm}{!}{\begin{tikzpicture}[
    >={Stealth[length=4pt,width=3.5pt]}, auto, semithick,
    every state/.style={draw=black!80,fill=black!3,minimum size=7.5mm,inner sep=0pt,font=\small},
    initial text={}, initial where=left, initial distance=4mm,
    accepting/.style={double distance=1.2pt},
    lbl/.style={font=\small,inner sep=1pt},
    every edge/.append style={draw=black!65}]
  \node[state,initial,fill=orange!25] (si)  at (0,0)       {$u_I$};
  \node[state] (s1)  at (2.0,1.0)  {$u_1$};
  \node[state] (s2)  at (3.7,1.0)  {$u_2$};
  \node[state] (s3)  at (5.4,1.8) {$u_3$};
  \node[state] (s5)  at (5.4,0.2) {$u_5$};
  \node[state] (s4)  at (7.0,1.0)  {$u_4$};
  \node[state] (s6)  at (1.3,-1.0)  {$u_6$};
  \node[state] (s7)  at (2.6,-1.0)  {$u_7$};
  \node[state] (s9)  at (3.9,-0.2) {$u_9$};
  \node[state] (s8)  at (3.9,-1.8) {$u_8$};
  \node[state] (s10) at (5.4,-1.0)  {$u_{10}$};
  \node[state] (s11) at (7.0,-1.0)  {$u_{11}$};
  \node[state,accepting,fill=green!12] (g) at (8.6,0) {$u_F$};

  \path[->]
    (si)  edge node[lbl] {$Y$} (s1)
    (si)  edge node[lbl,swap] {$B$} (s6)
    (s1)  edge node[lbl] {$R$} (s2)
    (s2)  edge[bend left=20]  node[lbl] {$Y$} (s3)
    (s3)  edge[bend left=20]  node[lbl,pos=0.22] {$\neg Y$} (s2)
    (s2)  edge[bend right=20] node[lbl,swap,pos=0.6] {$B$} (s5)
    (s5)  edge[bend right=20] node[lbl,swap,pos=0.22] {$\neg B$} (s2)
    (s3)  edge node[lbl] {$B$} (s4)
    (s5)  edge node[lbl] {$Y$} (s4)
    (s4)  edge node[lbl] {$T$} (g)
    (s6)  edge node[lbl,swap] {$R$} (s7)
    (s7)  edge[bend left=20]  node[lbl] {$B$} (s9)
    (s9)  edge[bend left=20]  node[lbl,pos=0.22] {$\neg B$} (s7)
    (s7)  edge[bend right=20] node[lbl,swap] {$Y$} (s8)
    (s8)  edge[bend right=20] node[lbl,swap,pos=0.22] {$\neg Y$} (s7)
    (s9)  edge node[lbl] {$Y$} (s10)
    (s8)  edge node[lbl,swap] {$B$} (s10)
    (s10) edge node[lbl,swap] {$P$} (s11)
    (s11) edge node[lbl,swap] {$T$} (g);
\end{tikzpicture}}}%
\def\ovContract#1#2#3{%
  \draw[ovCard] (5.15,#1-0.5) rectangle (8.85,#1+0.5);
  \node[anchor=west, font=\footnotesize\bfseries, text=ovSpec] at (5.2,#1+0.25) {$\mathcal{C}(#2)$};
  \node[anchor=west, font=\scriptsize\itshape, text=ovSpec!80!black] at (5.25,#1-0.06) {assume};
  \node[anchor=west, font=\scriptsize]                               at (6.65,#1-0.06) {$s_0 \in \mathcal{I}(u)$};
  \node[anchor=west, font=\scriptsize\itshape, text=ovSpec!80!black] at (5.25,#1-0.32) {guarantee};
  \node[anchor=west, font=\scriptsize]                               at (6.65,#1-0.32) {$\Pr(\tau_G \le t) \ge #3$};
}
\def\ovSlot#1#2#3{%
  \draw[ovImplBox] (14.35,#1-0.3) rectangle (16.55,#1+0.3);
  \node at (14.85,#1) {#3};
  \node[font=\scriptsize, anchor=west] at (15.2,#1) {$\pi(#2)$};
  \node[font=\small] at (16.8,#1) {$\models$};
  \draw[ovSpec, dashed, rounded corners=3pt, line width=0.6pt, fill=ovSpecFill!50] (17.05,#1-0.3) rectangle (18.25,#1+0.3);
  \node[font=\scriptsize, text=ovSpec] at (17.65,#1) {$\mathcal{C}(#2)$};
}
\def\ovpc#1{\scalebox{0.72}{#1}}%
\resizebox{\textwidth}{!}{%
\begin{tikzpicture}[font=\footnotesize,
  ovPanel/.style={draw=ovPanel, rounded corners=6pt, dashed, line width=0.4pt},
  ovTitle/.style={font=\small\bfseries, anchor=north},
  ovSub/.style={font=\scriptsize, text=black!55, align=center},
  ovArr/.style={-{Stealth[length=5pt]}, line width=0.7pt, draw=black!70},
  ovCard/.style={draw=ovSpec, fill=ovSpecFill, rounded corners=3pt, line width=0.6pt},
  ovImplBox/.style={draw=black!50, fill=ovImpl, rounded corners=4pt, line width=0.5pt},
  ovLBox/.style={ovImplBox, minimum width=3.6cm, align=center, inner sep=3pt, font=\scriptsize}]

\draw[ovPanel] (0,0) rectangle (4.6,5.6);
\node[ovTitle] at (2.3,5.5) {Heterogeneous team task};
\node[ovSub]   at (2.3,4.98) {encoded as an automaton.};
\node at (2.3,3.45) {\usebox{\ovMapBox}};
\node at (2.3,1.32) {\usebox{\ovAutBox}};
\node[font=\scriptsize] at (2.3,0.27) {require $\Pr(\text{reach } u_F) \ge p_{\mathrm{team}}$};

\draw[ovArr] (4.62,2.8) -- (4.98,2.8);

\draw[ovPanel] (5.0,0) rectangle (9.0,5.6);
\node[ovTitle] at (7.0,5.5) {Assume/Guarantee contracts};
\node[ovSub]   at (7.0,4.86) {formalize the task's decomposition\\into sub-tasks.};

\ovContract{3.55}{Y_1}{p_1}
\ovContract{2.4}{R_1}{p_2}
\ovContract{1.25}{T_1}{p_3}

\draw[ovArr] (9.02,2.8) -- (9.38,2.8);

\draw[ovPanel] (9.4,0) rectangle (13.8,5.6);
\node[ovTitle] at (11.6,5.5) {IHA-MARL};
\node[ovSub]   at (11.6,4.86) {for iterative task assignment\\and RL training.};

\node[ovLBox, minimum height=0.85cm] (assign) at (11.75,4.0)
  {{\footnotesize\bfseries Assign}\\[1pt]\ovpc{\pcKnight}\,$\mapsto Y_1$\; \ovpc{\pcRook}\,$\mapsto R_1$\; \ovpc{\pcKing}\,$\mapsto T_1$};
\node[ovLBox, minimum height=0.85cm] (train) at (11.75,2.85)
  {{\footnotesize\bfseries Train}\\RL policy per assigned contract};

\node[ovLBox, minimum height=1.9cm] (eval) at (11.75,1.15) {};
\node[font=\footnotesize\bfseries] at (11.75,1.88) {Evaluate};
\node[font=\scriptsize] at (11.75,1.62) {vs.\ required guarantee};
\def\ovbx{10.65}\def\ovbw{2.1}
\foreach \lab/\est/\req/\ok [count=\r from 0] in
    {Y_1/0.99/0.97/1, R_1/0.98/0.97/1, T_1/0.55/0.97/0} {
  \pgfmathsetmacro{\yy}{1.25-0.3*\r}
  \node[font=\tiny, anchor=east] at (\ovbx-0.03,\yy) {$\lab$};
  \fill[black!12] (\ovbx,\yy-0.07) rectangle ++(\ovbw,0.14);
  \fill[black!55] (\ovbx,\yy-0.07) rectangle ++(\est*\ovbw,0.14);
  \draw[ovSpec, line width=0.9pt] (\ovbx+\req*\ovbw,\yy-0.11) -- ++(0,0.22);
  \ifnum\ok=1
    \draw[ovOk, line width=0.8pt, line cap=round, line join=round]
      (\ovbx+\ovbw+0.12,\yy) -- ++(0.06,-0.07) -- ++(0.12,0.15);
  \else
    \draw[ovBad, line width=0.8pt, line cap=round]
      (\ovbx+\ovbw+0.13,\yy-0.07) -- ++(0.14,0.14) ++(-0.14,0) -- ++(0.14,-0.14);
  \fi
}
\node[font=\tiny, text=ovSpec, anchor=north, inner sep=1pt] at (\ovbx+0.97*\ovbw,0.53) {$p$};

\draw[ovArr] (assign) -- (train);
\draw[ovArr] (train) -- (eval);
\draw[ovArr] (eval.west) -- ++(-0.22,0) |- (assign.west)
   node[pos=0.25, rotate=90, anchor=south, font=\tiny, text=black!60, inner sep=2pt] {};

\draw[ovArr] (13.82,2.8) -- (14.18,2.8);

\draw[ovPanel] (14.2,0) rectangle (18.4,5.6);
\node[ovTitle] at (16.3,5.5) {Compose};
\node[ovSub]   at (16.3,4.98) {learned agent behaviors.};

\node[font=\tiny, text=black!55] at (15.45,4.5) {components};
\node[font=\tiny, text=black!55] at (17.65,4.5) {contracts};

\ovSlot{4.05}{Y_1}{\pcKnight}
\ovSlot{3.15}{R_1}{\pcRook}
\ovSlot{2.25}{T_1}{\pcKing}
\foreach \y in {3.6,2.7} {
  \node[font=\scriptsize] at (15.45,\y) {$\otimes$};
  \node[font=\scriptsize] at (17.65,\y) {$\otimes$};
}

\node[font=\small] at (16.3,1.5) {$\Downarrow$};
\node[draw=ovSpec, fill=ovSpecFill, rounded corners=4pt, line width=0.7pt, align=center, font=\scriptsize, inner sep=3pt]
  at (16.3,0.8) {team requirement met:\\[-1pt]$\Pr(\text{reach } u_F) \ge p_{\mathrm{team}}$};

\end{tikzpicture}%
}%
  \caption{
  \textit{Heterogeneous team task:} the team objective is encoded as an automaton over high-level events, together with a required probability $p_{\mathrm{team}}$ that the team reaches the accepting state $u_F$. 
  \textit{A/G contracts:} after indexing repeated events, the team-task automaton is automatically decomposed into agent-agnostic contracts, each specifying what the agent fulfilling the contract may assume and what its policy must guarantee.
  \textit{IHA-MARL:} we formulate and solve an optimization problem to assign contracts and required performance guarantees to agents with heterogeneous capabilities. 
  The assigned agents are trained with RL to satisfy their contracts, and the resulting policies are evaluated against the guarantees. 
  Assignments and contract requirements are iteratively re-optimized and training continues until every guarantee is met or the training budget is exhausted. 
  \textit{Compose:} After training, each learned component (an agent together with its policy) satisfies its contract.
  Because contract satisfaction is preserved under composition, the composed team behavior satisfies the overall team requirement.
  }
  \label{fig:overview}
\end{figure*}

To address these challenges, we develop a theory of \textit{assume/guarantee contracts for heterogeneous agents in MAS (AGC-MAS)}, a framework for expressing team tasks, decomposing them into subtasks, and assigning them to heterogeneous teams of agents.
While this theory applies generally to heterogeneous MAS, we focus on their development in the setting of multi-agent reinforcement learning (MARL), and we present \textit{Iterative Heterogeneous Assignment for MARL (IHA-MARL)}, an algorithm for task assignment and MARL in heterogeneous teams.
Taken together, AGC-MAS and IHA-MARL turn heterogeneous team tasks into modular learning problems that can be trained locally, reassigned as agent capabilities are learned, and composed to guarantee the desired team-level performance.

More specifically, AGC-MAS uses contracts to specify the requirements for individual subtasks so that, when those requirements are met, the team task is completed as a whole.
In particular, contracts define the conditions an agent may rely on while carrying out its subtask and the guarantees it must provide for that subtask to be considered safely completed.
Formally, each contract specifies a subtask, an automaton that tracks its high-level progress and coordination with other subtasks, timeout and success-probability requirements, and dispatch behavior specifying the order in which subtasks are attempted.

This contract structure localizes the learning and execution associated with each subtask to its assigned agent, replacing a monolithic full-team learning problem with smaller, well-defined, and more tractable learning problems. 
Its modularity allows an agent or policy to be replaced without changing the implementations of other contracts, while contract composition ensures that meeting the local requirements collectively guarantees the required team-level performance.
Our constructive algorithm automatically decomposes the team task into contracts with coordinating automata, while preserving the original team task under composition.
Finally, IHA-MARL learns during training which heterogeneous agents are suited to each contract and adjusts their assignments accordingly. 
It also adapts the contracts' required success probabilities to achieve the desired team-level success probability while minimizing the additional training required.

In a gridworld where agents with different movement capabilities coordinate pressure-plate activations to open gates, we demonstrate capability-aware reassignment and policy reuse.
Across three cooperative benchmarks, IHA-MARL achieves observed $100\%$ task success across five training seeds, with episode lengths comparable to the baselines. 
An empirical CPU timing pilot measured baseline training-loop times approximately $7$--$11\times$ those of IHA-MARL.

\begin{figure*}[t]
  \centering
  \begin{subfigure}[b]{0.40\textwidth}
    \centering
    \resizebox{!}{4.4cm}{\input{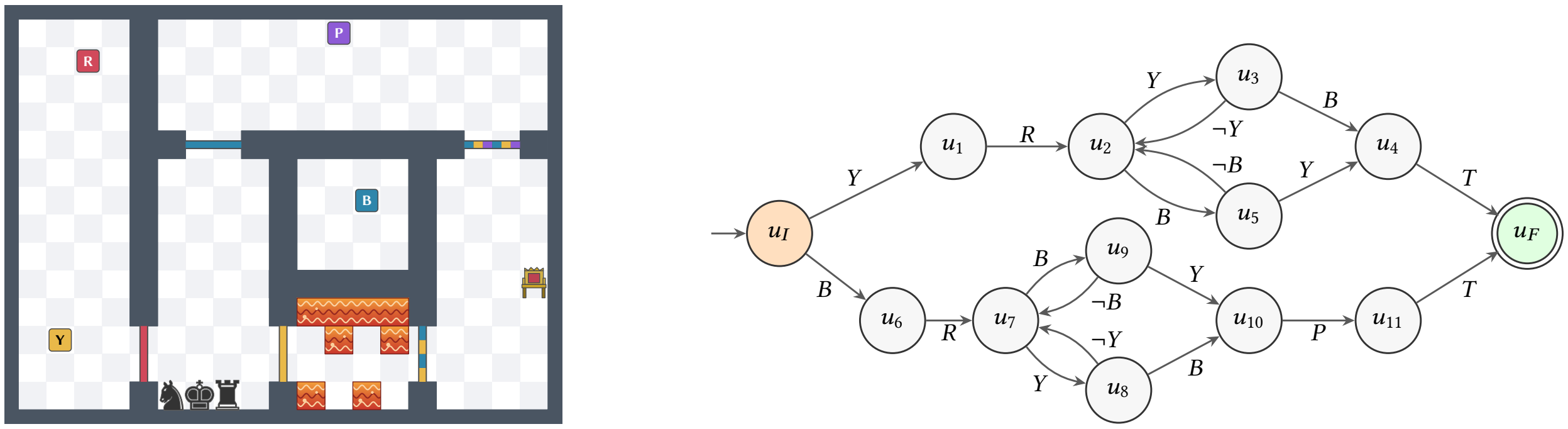}}
  \end{subfigure}\hfill
  \begin{subfigure}[b]{0.6\textwidth}
    \centering
    \resizebox{!}{4.0cm}{\begin{tikzpicture}[
    >={Stealth[length=4pt,width=3.5pt]}, auto, semithick,
    every state/.style={draw=black!80,fill=black!3,minimum size=7.5mm,inner sep=0pt,font=\small},
    initial text={}, initial where=left, initial distance=4mm,
    accepting/.style={double distance=1.2pt},
    lbl/.style={font=\small,inner sep=1pt},
    every edge/.append style={draw=black!65}]
  \node[state,initial,fill=orange!25] (si)  at (0,0)       {$u_I$};
  \node[state] (s1)  at (2.0,1.0)  {$u_1$};
  \node[state] (s2)  at (3.7,1.0)  {$u_2$};
  \node[state] (s3)  at (5.4,1.8) {$u_3$};
  \node[state] (s5)  at (5.4,0.2) {$u_5$};
  \node[state] (s4)  at (7.0,1.0)  {$u_4$};
  \node[state] (s6)  at (1.3,-1.0)  {$u_6$};
  \node[state] (s7)  at (2.6,-1.0)  {$u_7$};
  \node[state] (s9)  at (3.9,-0.2) {$u_9$};
  \node[state] (s8)  at (3.9,-1.8) {$u_8$};
  \node[state] (s10) at (5.4,-1.0)  {$u_{10}$};
  \node[state] (s11) at (7.0,-1.0)  {$u_{11}$};
  \node[state,accepting,fill=green!12] (g) at (8.6,0) {$u_F$};

  \path[->]
    (si)  edge node[lbl] {$Y$} (s1)
    (si)  edge node[lbl,swap] {$B$} (s6)
    (s1)  edge node[lbl] {$R$} (s2)
    (s2)  edge[bend left=20]  node[lbl] {$Y$} (s3)
    (s3)  edge[bend left=20]  node[lbl,pos=0.22] {$\neg Y$} (s2)
    (s2)  edge[bend right=20] node[lbl,swap,pos=0.6] {$B$} (s5)
    (s5)  edge[bend right=20] node[lbl,swap,pos=0.22] {$\neg B$} (s2)
    (s3)  edge node[lbl] {$B$} (s4)
    (s5)  edge node[lbl] {$Y$} (s4)
    (s4)  edge node[lbl] {$T$} (g)
    (s6)  edge node[lbl,swap] {$R$} (s7)
    (s7)  edge[bend left=20]  node[lbl] {$B$} (s9)
    (s9)  edge[bend left=20]  node[lbl,pos=0.22] {$\neg B$} (s7)
    (s7)  edge[bend right=20] node[lbl,swap] {$Y$} (s8)
    (s8)  edge[bend right=20] node[lbl,swap,pos=0.22] {$\neg Y$} (s7)
    (s9)  edge node[lbl] {$Y$} (s10)
    (s8)  edge node[lbl,swap] {$B$} (s10)
    (s10) edge node[lbl,swap] {$P$} (s11)
    (s11) edge node[lbl,swap] {$T$} (g);
\end{tikzpicture}}
  \end{subfigure}
  \caption{Left: An illustration of the Chess-Maze environment. A knight \pcKnight, king \pcKing, and rook \pcRook{} must learn to cooperate in order to press the colored pressure plates and open the gates of matching color so that the king may reach the throne. Multicolored gates open only while all of their plates are pressed simultaneously. Right: An automaton encoding this team task. Each transition is labeled by an event: $Y$, $R$, $B$, and $P$ denote pressing the corresponding pressure plate, $\neg Y$ and $\neg B$ denote releasing it, and $T$ denotes the king reaching the throne. The upper and lower branches are alternative plans for completing the task. 
  }
  \label{fig:chessmaze}
\end{figure*}

\section{Related Works}

Distributed MARL leverages the decentralized structure of multi-agent tasks to enable parallel learning, robustness, and scalability \cite{MARL_SURVEY}. 
Automata, long used in formal methods and model checking to specify and reason about system behavior, provide one way to structure this decentralization by representing high-level task progress and coordination \cite{Principles_of_Model_Checking,LTL_MARL,OG_Product_MDP}.
More recently, reinforcement-learning frameworks such as Reward Machines (RMs) have used automata as high-level semantic descriptions of tasks, in which automaton states track task progress and transitions correspond to meaningful subgoals whose completion can provide reward to low-level policies \cite{OG_RM_Paper}. 
This idea has also been extended to multi-agent settings, where local automata distribute high-level task information among agents and synchronize their progress toward a shared objective \cite{RM_paper}. 

Existing approaches to decomposing such team-task automata construct local automata through projections onto subsets of the event alphabet and characterize when their composition preserves the original task \cite{AutomataTaskDecomposition}. In this work, we distinguish two roles that events play in these local automata: an agent may be responsible for realizing an event, or may need only to observe that event in order to synchronize its progress with the rest of the team. Projection at the level of event type does not distinguish different occurrences of the same event within a task, thereby coupling these two roles across all such occurrences. We instead index individual event occurrences, allowing realization and synchronization requirements to be specified independently at each point in the team task. This finer-grained representation enables the constructive decomposition developed in this paper.

Assume/guarantee (A/G) contract theory separates component requirements from their implementations and supports compositional reasoning about system behavior \cite{contracts_for_system_design}. 
A/G formulations have been applied in multi-agent and dynamical-system settings \cite{assume_guarantee_contracts_for_dynamical_systems,AG_STL_MAS}, and assumption/guarantee ideas have also been introduced in reinforcement learning \cite{assume_guarantee_reinforcement_learning}. 
However, existing RL formulations do not, to our knowledge, retain the full compositional structure of contract theory, which we adapt to MARL to support modular reasoning about learned agent behaviors. 

On the algorithmic side, IHA-MARL builds on the training-sensitive optimization framework of \citet{VCRL}, extending it to jointly determine heterogeneous agent--subtask assignments during training.

\section{Method}

As a running example, we introduce Chess-Maze (Figure~\ref{fig:chessmaze}), where a rook, king, and knight activate pressure plates to open gates and let the king reach the throne. Some gates require several plates to be held simultaneously. An automaton tracks the required activations and releases.

\subsection{Multi-Agent Reinforcement Learning}
\label{subsec:multi_agent_setting}
We model the underlying multi-agent MDP dynamics by
\(\mathcal{M}=(S,A,P)\),
with the team task specified and learning objectives derived below.
Let
\(
I = \{i_1,i_2,\ldots,i_n\}
\)
denote the set of agents.
The joint state space is
\(
S = S_1 \times \cdots \times S_n \times S_e,
\)
where $S_m$ denotes the state space of agent $i_m$, for $1 \leq m \leq n$, and $S_e$ denotes the state space of the environment.
Let $s_{\mathrm{init}}\in S$ denote the initial joint state of the multi-agent environment.
Similarly, the joint action space is
\(
A = A_1 \times \cdots \times A_n,
\)
where $A_m$ is the action space of agent $i_m$.
The transition kernel
\(
P(s' \mid s,a)
\)
defines a probability distribution over $S$ for each $s \in S$ and $a \in A$.

For a subset of agents $D = \{i_{j_1}, \ldots, i_{j_k}\}$, let $S_D$ be the state space $S_{j_1} \times \ldots \times S_{j_k}$.
For $s=(s_1,\ldots,s_n,s_e)$, its coordinate projection onto $S_D$ is $s_D=(s_{j_1},\ldots,s_{j_k})$, retaining only the tuple coordinates of agents in $D$; $s_{-D}$ retains those of the complement agents $I\setminus D$.
For a single agent $i_j$, we write $s_{-j}$ analogously.
We use the same indexing convention for actions.

We do not require the dynamics of the agents to evolve independently in general.
Instead, we assume that each agent $i_j$ admits an \emph{idle policy}
\(
\pi_{\mathrm{idle}}^j : S_j \rightarrow \Delta(A_j),
\)
under which the agent does not interfere with the dynamics of agents participating in the current task, where $\Delta(A_j)$ denotes the set of probability distributions over $A_j$.
More precisely, for any subset $D \subseteq I$, if every agent $i_j \notin D$ follows its idle policy, so that
\(
a_j \sim \pi_{\mathrm{idle}}^j(s_j),
\textrm{ for } i_j \notin D,
\)
then the transition dynamics factor as
\[
\begin{aligned}
&P(s_D',s_{-D}',s_e'
    \mid
    s_D,s_{-D},s_e,a_D,a_{-D})
\\
&\qquad =
P_D(s_D',s_e' \mid s_D,s_e,a_D)
P_{-D}(s_{-D}' \mid s_{-D},a_{-D}),
\end{aligned}
\]
for appropriate transition kernels $P_D$ and $P_{-D}$.
Thus, when the agents outside $D$ follow their idle policies, they neither affect the environment state nor interfere with the transition dynamics of the agents in $D$.

In Chess-Maze, each agent state records its piece's grid position,
while $S_e$ records pressure-plate and gate states.
Actions consist of the movements permitted by the corresponding chess
piece together with a no-op; attempted moves through walls or closed
gates leave the agent in place, while pressure-plate occupancy updates
the gate states.
Agents may share cells, so an idle agent repeatedly selects the no-op,
preserving any plate it already occupies without obstructing other agents.

\subsection{Automata as Descriptions of Team Tasks}
We use automata to encode high-level task progress \cite{RM_paper, LOTaD}: events represent subgoal completions, low-level policies realize these events, and the automaton coordinates their composition toward the overall task.
\begin{definition}
An automaton is a tuple $A = (\Sigma, U, u_I, F, \to)$ where $\Sigma$ is a finite set of events, $U$ is a finite set of states, $\to \subset  U \times \Sigma \times  U$ is a transition relation, $u_I$ is an initial state, and $F$ is a set of final states indicating task completion.
For brevity, we write $(u, \alpha, v) \in\to$ as $u \to_\alpha v$.
An automaton is \textit{deterministic} if $u \to_{\alpha} v$ and $u \to_{\alpha} w$ implies $v = w$.
\end{definition}

In Chess-Maze, the automaton in Figure~\ref{fig:chessmaze} describes the high-level task of activating the pressure plates in a sequence that opens the gates and allows the king to reach the throne.
For example, the event $B$ represents activation of the blue pressure plate.
Some gates require several pressure plates to be activated before they open.
As the required events occur, the automaton transitions through states representing the team's progress toward task completion.

Individual agents, however, need not have access to the state of the complete team-task automaton or be aware of every event that occurs.
An agent may instead maintain a local automaton describing only the events and task progress relevant to its role.
Some events are realized by the agent itself, while others are realized by teammates and observed through communication, allowing the agent to track task progress and coordinate when to act.
For example, rather than tracking the state of the complete team-task automaton, the king may only need to track whether the knight or rook has opened the final gate, an event it observes but does not realize itself. Once this event occurs, the king can proceed to reach the throne, an event that it realizes directly.

When an event is relevant to multiple agents, their local automata synchronize on that shared event.
The king synchronizes with the knight on the event of the final gate opening, causing both local automata to advance to states in which the remaining task is for the king to reach the throne.
Parallel composition provides a formal mechanism for combining such local automata while preserving this synchronization.

\begin{definition} (Parallel Composition of Automata)
The parallel composition of automata $A_1 = (\Sigma_1, U_1, u_{I}^1, F_1, \to_1)$ and $ A_2 = (\Sigma_2, U_2, u_{I}^2, F_2, \to_2)$ we define as $A_1 \parallel A_2 = (\Sigma, U, u_I, F, \to),$ where $\Sigma = \Sigma_1 \cup \Sigma_2$, $U = U_1 \times U_2$, $u_I = (u_I^1, u_I^2)$, $F = F_1 \times F_2$, and 
\begin{align*}
& (u_1, u_2) \to_\alpha (u_1', u_2') \in \to \iff \\
&\quad \begin{cases}
    u_1 \to_\alpha u_1' \wedge u_2 \to_\alpha u_2' & \text{ if } \alpha \in \Sigma_1 \cap \Sigma_2\\
    u_1 \to_\alpha u_1' \wedge u_2 = u_2' & \text{ if } \alpha \in \Sigma_1 \setminus \Sigma_2\\
    u_2 \to_\alpha u_2' \wedge u_1 = u_1' & \text{ if } \alpha \in \Sigma_2 \setminus \Sigma_1\\
    \text{undefined} & \text{else}
\end{cases}. 
\end{align*}
\end{definition}
The operation of parallel composition is associative \cite{Principles_of_Model_Checking}, so we lose no sleep in defining the composition of multiple automata as
\[
    \parallel_{i = 1}^n A_i = A_1 \parallel ( A_2 \parallel \dots (A_{n-1} \parallel A_n)\dots ).
\]

\subsection{Linking Automata to the Multi-Agent Environment}
\label{subsec:labeling_function}
In order to make use of the automata for RL tasks, we need to create a bridge between high-level events and the low-level dynamics.
To do so, we use labeling functions \cite{RM_paper}.
Let
\(
L : U \times S \to 2^{\Sigma}
\)
map each low-level state at a task state to the events that hold at that state.
From this map, we can derive a low-level goal set, into which a policy must drive the system in order to realize an event.
If we would like to trigger event $\alpha$ at state $u$, to effect the transition $u \to_{\alpha} u'$, then the goal set is $G(u, \alpha) = \{s \in S : \alpha \in L(u, s)\}$.

To train such a policy, we must also determine the low-level entry set from which the policy may begin when the automaton enters state $u$. Let $\mathrm{prec}(u)\subseteq\to$ be the set of transitions with destination $u$. The entry set includes their goal sets and, at $u_I$, the initial joint state:
\[
\mathcal{I}(u) = \bigcup_{(v, \alpha, u) \in \mathrm{prec}(u)} G(v, \alpha)
\;\cup\;
\begin{cases}
\{s_{\mathrm{init}}\}, & u=u_I,\\
\varnothing, & u\neq u_I.
\end{cases}
\]

Therefore, in order for a low-level policy to trigger a high level event $\alpha$ at high level state $u$, it must be trained to produce actions that bring the system from any state in $\mathcal{I}(u)$ to a state in $G(u, \alpha)$.
For example, a policy trained on the knight to press the blue button must find a path from any initial knight position with the blue gate yet unopened to a state where the knight is on the blue pressure plate.

\subsection{Indexed Automata and Task Decomposition}

In many automata-based multi-agent formulations, events are distributed among agents at the level of event type~\cite{RM_paper,LOTaD}.
For example, if the knight's local automaton contains the yellow pressure-plate event, then every occurrence of that event is represented in its local task, even if only some occurrences are relevant to the knight.
However, when the same event occurs at multiple points in the team task, it may be desirable for an agent to realize the event or synchronize with another agent's realization of it on only particular occurrences.

In Chess-Maze, for example, the yellow pressure plate must be activated twice.
On its first occurrence, the knight activates it to open the yellow gate, since before the red gate is opened the knight is the only agent capable of reaching the plate by jumping over the walls.
On its second occurrence, however, the rook must activate the yellow plate while the knight simultaneously occupies the blue plate to open the final blue/yellow gate.
The rook has no need to observe or synchronize on the first yellow-plate event.
Requiring every occurrence of the same event type to be treated identically therefore imposes an unnecessary restriction on task allocation and synchronization, preventing different occurrences from being distributed independently among agents.

To distinguish these occurrences, we index events with the source state of their transition.
Thus, the two yellow-plate events may be represented as $Y_1$ and $Y_2$.
Although both correspond to the same underlying event, their indices distinguish their roles within the team task and allow them to be assigned independently to different agents.
This motivates the following definition.

\begin{definition}(Indexed Automaton)
Let $A=(\Sigma,U,u_I,F,\to)$
be a deterministic automaton.
The \emph{indexed automaton} of $A$ is the automaton $A'=(\Sigma',U,u_I,F,\to'),$
where
\vspace{-2mm}\[
\Sigma'
=
\{(u,\alpha)\in U\times\Sigma :
\exists v\in U \text{ such that } u\to_\alpha v\}.
\vspace{-2mm}
\]
For notational convenience, we write the indexed event
$(u,\alpha)$ as $\alpha_u$.
Define $u\to'_{\alpha_u}v$ iff $u\to_\alpha v$.
\end{definition}
We call an automaton \emph{indexed} if it is obtained by this indexing construction.

When the team-task automaton is specified first, we must derive local automata whose parallel composition preserves it, so that completing the local subtasks guarantees completion of the original task.
We formalize this requirement using automaton isomorphism, a stronger relation than the bisimulation notions commonly used in the literature.

Let $\Sigma^*$ denote the set of finite event strings over $\Sigma$.
For $\sigma=\alpha_1\cdots\alpha_n\in\Sigma^*$, write $u\xrightarrow{\sigma}v$ when there is a path $u\xrightarrow{\alpha_1}\cdots\xrightarrow{\alpha_n}v$.
A state $u$ is \emph{reachable} if $u_I\xrightarrow{\sigma}u$ for some $\sigma\in\Sigma^*$.
Define
\[
\begin{aligned}
r(u) &= \{v\in U:\exists\sigma\in\Sigma^*,\;
u\xrightarrow{\sigma}v\},\\
s(u) &= \{v\in U:\exists\sigma\in\Sigma^*,\;
v\xrightarrow{\sigma}u\}\setminus\{u\}.
\end{aligned}
\]
Thus, $r(u)$ contains the states reachable from $u$, while $s(u)$
contains the states from which $u$ is reachable, excluding $u$ itself.
\begin{definition}(Isomorphic Automata)
Let $A_i = (\Sigma, U_i, u_{I}^i, F_i, \to_i)$, $i = 1, 2$ be automata.
We say that $A_1$ and $A_2$ are \textit{isomorphic} if there exists a bijection $f : r(u_I^1) \to r(u_I^2)$ such that $f(u_I^1)=u_I^2$ and, for all $u_1,u_1'\in r(u_I^1)$ and $\alpha\in\Sigma$, $u_1\to_\alpha^1 u_1'$ if and only if $f(u_1)\to_\alpha^2 f(u_1')$; moreover, for every $u\in r(u_I^1)$, $u\in F_1$ if and only if $f(u)\in F_2$.
Then, we write $A_1 \cong A_2$.
\end{definition}

Given a team-task automaton $A$, we now ask how it can be decomposed into smaller automata whose parallel composition preserves the original task.
Existing approaches construct agent automata by projecting the team-task automaton onto candidate subsets of its event alphabet and then verify whether their parallel composition preserves the original task~\cite{AutomataTaskDecomposition,auto_decompose_RM}.
Finding a desirable decomposition may therefore require searching over different choices of local event sets.
By indexing individual event occurrences, we instead obtain the following constructive decomposition with a fixed three-state structure.

\begin{theorem}
\label{thm:big_thm}
Let $A = (\Sigma, U, u_I, F, \to)$ be a deterministic indexed automaton with a unique sink state $s \in F$, such that every state is reachable from $u_I$ and has a path to $s$.
Then, there exist $|\Sigma|$ automata $A_i = (\Sigma_i, U_i, u_I^i, F_i, \to_i)$ with $U_i = \{\mathrm{armed}, \mathrm{active}, \mathrm{resolved}\}$ and $\bigcup_i \Sigma_i = \Sigma$ such that $A \cong \parallel_i A_i$.
\end{theorem}

The proof of Theorem~\ref{thm:big_thm} is constructive. We give the
construction here, since it also provides the intuition behind the
resulting local automata.

Let $\Sigma=\{\alpha_1,\ldots,\alpha_m\}$, where each indexed event
$\alpha_i$ labels a unique transition $u_i\to_{\alpha_i}v_i$.
For each $\alpha_i$, partition the team-task states into
\(U^i_{\mathrm{armed}}=s(u_i)\),
\(U^i_{\mathrm{active}}=\{u_i\}\), and
\(U^i_{\mathrm{resolved}}
=U\setminus\bigl(s(u_i)\cup\{u_i\}\bigr)\).
Let
\(q_i:U\to\{\mathrm{armed},\mathrm{active},\mathrm{resolved}\}\)
map each team-task state to its corresponding partition element.

The local automaton associated with $\alpha_i$ retains precisely the
indexed events needed to update this three-state view. Specifically,
define
\(\Sigma_i=
\{\alpha_j\in\Sigma:q_i(u_j)\neq q_i(v_j)\}\cup\{\alpha_i\}\),
and let
\(A_i=(\Sigma_i,U_i,u_I^i,F_i,\to_i)\), where
\(U_i=\{\mathrm{armed},\mathrm{active},\mathrm{resolved}\}\),
\(u_I^i=q_i(u_I)\),
\(F_i=\{q_i(u):u\in F\}\), and, for every
$\alpha_j\in\Sigma_i$,
\(q_i(u_j)\to_{\alpha_j}^i q_i(v_j)\).
Since indexed events label unique transitions, each $A_i$ is
deterministic, and \(\bigcup_i\Sigma_i=\Sigma\).

Each $A_i$ therefore stores exactly the high-level information needed to track its indexed event $\alpha_i$: whether that event may be required in the future (\emph{armed}), is enabled at the current team-task state (\emph{active}), or can no longer be required (\emph{resolved}). 
Events other than $\alpha_i$ appear in its alphabet only when their occurrence changes this status, allowing the local automata to remain synchronized with the evolving team task. 
We call $A_i$ a \emph{compositional atomic subtask automaton} (CASA): it is \emph{atomic} because it is centered on the realization of a single indexed event, and \emph{compositional} because
Theorem~\ref{thm:big_thm} establishes that the parallel composition of all such CASAs reconstructs the original team-task automaton. 
The complete proof constructs the corresponding isomorphism and is provided in the supplementary material.

\begin{definition}
We call the automaton $A_i$ constructed above for indexed event $\alpha_i$ its \emph{compositional atomic subtask automaton} (CASA).
\end{definition}
We also write $A_{\alpha_u}=A_i$ when $\alpha_u=\alpha_i$.

\subsection{Assume/Guarantee Contract Theory}

We assume that indexed events in the team-task automaton are realized sequentially: at any time, at most one indexed event is dispatched for implementation.
After an event is realized and the team-task automaton reaches its next state, several successor events may be available.
A choice must therefore be made as to which event to \emph{dispatch} for realization next.
For example, at the initial state of Chess-Maze, the team must choose between alternative paths through the gates at the initial automaton state.
We represent these choices using a high-level \emph{dispatching policy}.

For an indexed event $\alpha_u$ corresponding to the transition $u\to_\alpha v$, define $\mathrm{succ}(\alpha_u)$ to be the set of indexed events whose transitions leave $v$. Similarly, $\mathrm{succ}(u_I)$ is the set of indexed events whose transitions leave the initial state.

\begin{definition}[Dispatching Policy]
A \emph{dispatching policy} is a function $\mu:(\{u_I\}\cup\Sigma)\times\Sigma\to[0,1]$ such that $\mu(\beta\mid x)=0$ whenever $\beta\notin\mathrm{succ}(x)$ and
\(
\sum_{\beta\in\mathrm{succ}(x)}\mu(\beta\mid x)=1
\)
whenever $\mathrm{succ}(x)\neq\emptyset$.
\end{definition}

At initialization, the first indexed event is sampled according to $\mu(\cdot\mid u_I)$.
The initial dispatch policy is implemented by an external agent that commissions the team.
Thereafter, when an indexed event $\alpha_u$ is successfully realized, the next event is sampled according to $\mu(\cdot\mid\alpha_u)$. 
Thus, $\mu$ acts as a high-level policy over indexed events.

Let $p_{\mathrm{team}}\in[0,1]$ denote the desired probability that the team automaton reaches a state in $F$ and thereby completes the task successfully.
To support such a team-level requirement, we associate each indexed event $\alpha_u$ with a timeout $t(\alpha_u)>0$ and a required success probability $p(\alpha_u)\in[0,1]$.
These event-level quantities form the performance requirements that will later be assigned through contracts.

We call any subset $\mathcal{X}\subseteq\Sigma$ a \emph{subtask}. A contract specifies a subtask together with requirements on its synchronization, reliability, timeout, and dispatch behavior.

\begin{definition}[Contract]
A \emph{contract} is a tuple
\[
\mathcal{C}=(\mathcal{X}_{\mathcal{C}},\mathcal{A}_{\mathcal{C}},t_{\mathcal{C}},p_{\mathcal{C}},\mu_{\mathcal{C}}),
\]
where $\varnothing\neq\mathcal{X}_{\mathcal{C}}\subseteq\Sigma$ is a subtask,
$\mathcal{A}_{\mathcal{C}}=\parallel_{\alpha_u\in\mathcal{X}_{\mathcal{C}}}A_{\alpha_u}$ is the parallel composition of the CASAs corresponding to its indexed events,
$t_{\mathcal{C}}:\mathcal{X}_{\mathcal{C}}\to\mathbb{R}_{>0}$ specifies the timeout for each indexed event,
$p_{\mathcal{C}}:\mathcal{X}_{\mathcal{C}}\to[0,1]$ specifies its required probability of successful realization, and
$\mu_{\mathcal{C}}=\mu|_{\mathcal{X}_{\mathcal{C}}\times\Sigma}$ specifies its dispatch behavior.
\end{definition}

A contract specifies the requirements that a subtask must satisfy, while a component specifies how that subtask is implemented.
A component assigns a subteam of agents to each event in the subtask and specifies the policy executed by that subteam.
Not every such assignment is feasible.
For example, in Chess-Maze, the knight cannot be assigned to realize both $Y_2$ and $B_2$, since the two plates must be pressed simultaneously to open the gate.

Let $\mathfrak{D}\subseteq(2^I)^\Sigma$ denote the set of legal assignments of indexed events to subteams.
We call $D:\mathcal{X}\to2^I$ a legal sub-assignment if it is the restriction of some $d\in\mathfrak{D}$.
Let $\Delta(A_D)$ denote the set of probability distributions over the joint action set $A_D$.

\begin{definition}[Component]
A \emph{component} is a tuple
\[
M=(\mathcal{X}_M,D_M,\Pi_M,\mu_M),
\]
where $\mathcal{X}_M\subseteq\Sigma$ is the set of indexed events implemented by the component, $D_M:\mathcal{X}_M\to2^I$ is a legal sub-assignment, for each $\alpha_u\in\mathcal{X}_M$,
$\Pi_M(\alpha_u):S_{D_M(\alpha_u)}\times S_e\to\Delta(A_{D_M(\alpha_u)})$
is a policy used by the assigned agents to realize $\alpha_u$, and
$\mu_M:\mathcal{X}_M\times\Sigma\to[0,1]$
is the component's dispatching policy.
\end{definition}

When $\alpha_u$ is dispatched, the agents in $D_M(\alpha_u)$ execute $\Pi_M(\alpha_u)$ while all other agents follow their idle policies. By the idle-policy assumption, these agents do not interfere with the dynamics of the dispatched subteam.

The low-level objective of each policy is derived from the labeling function defined in Subsection~\ref{subsec:labeling_function}.
For $D\subseteq I$, define the \emph{local goal set}
\(
G_D(\alpha_u)=\{(s_D,s_e):\alpha\in L(u,(s_D,s_{-D},s_e))\ \text{for all }s_{-D}\in S_{-D}\},
\)
using the state projection notation defined in Subsection~\ref{subsec:multi_agent_setting}.
Thus, reaching $G_D(\alpha_u)$ guarantees that $\alpha$ is realized independently of the states of agents outside $D$.

Likewise, define the \emph{local entry set}
\[
\mathcal{I}_D(u)=\{(s_D,s_e):\exists s_{-D}\text{ such that }(s_D,s_{-D},s_e)\in\mathcal{I}(u)\}.
\]
This is the projection of the possible low-level states encountered when the team-task automaton enters $u$ onto the state coordinates of the participating agents.

The subteam assigned to a dispatched event reports whether it successfully realizes that event; undispatched events are ignored by the automaton.
For $\alpha_u\in\mathcal{X}_M$, let
$\pi(\alpha_u)=(\Pi_M(\alpha_u),\pi_{\mathrm{idle}}^{-D_M(\alpha_u)})$
denote the joint policy under which the assigned subteam executes $\Pi_M(\alpha_u)$ and all other agents execute their idle policies. For a set $E$, let $\tau_E$ denote its first hitting time.
The success probability achieved by component $M$ on $\alpha_u$ within time $t$ is
\begin{equation}
\label{eq:component_success}
p_M(\alpha_u;t)
=
\inf_{s^0\in\mathcal{I}_{D_M(\alpha_u)}(u)}
\Pr\!\left(
\tau_{G_{D_M(\alpha_u)}(\alpha_u)}\leq t
\mid s^0,\pi(\alpha_u)
\right).
\end{equation}

\begin{definition}[Contract Satisfaction]
Component $M$ \emph{satisfies} contract $\mathcal{C}$, written $M\models\mathcal{C}$, if $\mathcal{X}_{\mathcal{C}}\subseteq\mathcal{X}_M$ and, for every $\alpha_u\in\mathcal{X}_{\mathcal{C}}$,
\(
p_M(\alpha_u;t_{\mathcal{C}}(\alpha_u))
\geq
p_{\mathcal{C}}(\alpha_u),
\)
and
\(
\mu_M(\beta\mid\alpha_u)
=
\mu_{\mathcal{C}}(\beta\mid\alpha_u)
\text{ for all }\beta\in\Sigma.
\)
\end{definition}

\paragraph{Composition.}
Two contracts $\mathcal{C}_1$ and $\mathcal{C}_2$ are \emph{composable} if $\mathcal{X}_{\mathcal{C}_1}\cap\mathcal{X}_{\mathcal{C}_2}=\emptyset$. Their composition is
\[
\mathcal{C}_1\otimes\mathcal{C}_2
=
\bigl(\mathcal{X}_{\mathcal{C}_1}\cup\mathcal{X}_{\mathcal{C}_2},
\mathcal{A}_{\mathcal{C}_1}\parallel\mathcal{A}_{\mathcal{C}_2},
t_{\mathcal{C}_1}\cup t_{\mathcal{C}_2},
p_{\mathcal{C}_1}\cup p_{\mathcal{C}_2},
\mu_{\mathcal{C}_1}\cup\mu_{\mathcal{C}_2}\bigr).
\]
Because their event sets are disjoint, these unions are well-defined. The automaton component is determined by the combined subtask through the CASA construction, up to the canonical identification of product states.

Similarly, components $M_1$ and $M_2$ are composable if $\mathcal{X}_{M_1}\cap\mathcal{X}_{M_2}=\emptyset$ and $D_{M_1}\cup D_{M_2}$ is a legal sub-assignment. Their composition is obtained by taking the corresponding unions of their event sets, assignments, and policies.

\begin{proposition}
\label{prop:composition}
If $M_1\models\mathcal{C}_1$ and $M_2\models\mathcal{C}_2$, and both pairs are composable, then
\(
M_1\otimes M_2\models\mathcal{C}_1\otimes\mathcal{C}_2.
\)
\end{proposition}

This follows immediately because the requirements of a contract are event-wise, and composition preserves the satisfying implementation of each event.

For convenience, we call a contract $\mathcal{C}$ \emph{atomic} if $|\mathcal{X}_{\mathcal{C}}|=1$, and likewise call a component $M$ \emph{atomic} if $|\mathcal{X}_M|=1$.
Every contract can be expressed as the composition of atomic contracts, and every component as the composition of atomic components.
\paragraph{Closed Contracts.}
A contract $\mathcal{C}$ is \emph{closed} if $\mu(\alpha_u\mid u_I)=0$ for every $\alpha_u\notin\mathcal{X}_{\mathcal{C}}$ and $\mu_{\mathcal{C}}(\beta_v\mid\alpha_u)=0$ for every $\alpha_u\in\mathcal{X}_{\mathcal{C}}$ and $\beta_v\notin\mathcal{X}_{\mathcal{C}}$.
Thus, the initial dispatch lies within the contract and any subsequent dispatch after an event implemented by the contract also lies within the contract.
A closed contract therefore specifies a self-contained high-level task whose dispatched execution requires no external event implementation.

The contracts, components, and their composition operations instantiate the contract-theoretic framework of \citet{contracts_for_system_design}.
By separating the requirements of a subtask from its implementation, this formulation allows the team to replace one satisfying component with another without reconsidering the correctness of the overall team task.
It also enables distributed learning and execution, since the policies associated with different subtasks can be trained and executed by their assigned subteams rather than by the full team jointly.
Finally, the modular structure of contracts enables components to be reused across different higher-level contracts.
For example, a component in which the knight activates the initial yellow pressure plate can be reused regardless of whether the subsequent team-level dispatching policy sends the team along the upper or lower path to the throne.

\subsection{Iterative Heterogeneous Assignment}
Given a desired team completion probability $p_{\mathrm{team}}$, we seek a collection of atomic contracts whose composition is closed and whose collective satisfaction guarantees the team-level requirement. We may restrict attention to atomic contracts because every contract can be expressed as a composition of them. These contracts specify the performance required for each indexed event, but do not assume that satisfying components are already available.

Here, IHA-MARL uses single-agent assignments, while the contract framework supports extensions to subteams. To construct a component for each atomic contract, we must determine which agent should be assigned to its indexed event and train a policy for that agent to realize the event with the required performance. Thus, we must jointly determine the agent assignments and the policies to train on those assignments. These decisions are coupled, since suitable assignments depend on the capabilities of the available policies, while those capabilities are themselves learned through training.

IHA-MARL addresses this coupling iteratively. 
At each iteration, the optimization chooses atomic contract requirements that collectively guarantee the team-level requirement while remaining as close as possible to the current performance of the assigned policies, thereby minimizing the additional training required.
The policies associated with the selected assignments are then trained further, their performance is re-evaluated, and the optimization is solved again.

Since each atomic contract contains a single indexed event, we identify it by its event $x\in\Sigma$. 
A pair $(i,x)$ denotes the candidate assignment of agent $i$ to event $x$, for which we maintain a policy trained to realize $x$. 
Let $z_{i,x}\in\{0,1\}$ indicate whether this assignment is selected, and let $\sigma_{i,x}\in[0,1]$ denote the current estimated success probability of its policy, evaluated according to Equation~\ref{eq:component_success} using the fixed timeout $t_x$. 
In practice, $\sigma_{i,x}$ is estimated empirically from rollouts initialized from the corresponding entry set and is set to zero if no policy has yet been trained for $(i,x)$.
These empirical estimates are not certified lower bounds on the worst-case success probabilities.
Following \cite{VCRL}, Hoeffding's inequality can quantify uncertainty in success estimates from independent evaluation rollouts. Certification of the worst-case probability in Equation~\ref{eq:component_success} additionally requires bounds that hold uniformly over the entry set.

For each $x\in\Sigma$, let $p_x\in[0,1]$ denote the required success probability of its atomic contract and let $d_x\geq0$ denote its discounted occupancy in the high-level dispatch process. 
Together with the corresponding dispatch behavior, a selected assignment and its policy define an atomic component.

For the optimization, we collapse the accepting states into the terminal sink state of the automaton $u_F$ and assign reward $r_x=1$ to indexed events whose transitions enter $u_F$ through the collapsed accepting states, with $r_x=0$ otherwise. 
The resulting expected discounted reward therefore represents discounted task-completion probability. 
We add an artificial terminal self-loop $u_F\to_\epsilon u_F$, with $p_\epsilon=1$ and $r_\epsilon=0$, for flow bookkeeping. It acts as a flow sink: its occupancy absorbs flow reaching $u_F$ without contributing to the incoming-flow sums.
The flow constraints below use the state set and transition incidences of this modified automaton.
Let $R^*$ denote the maximum discounted completion reward obtainable under the same flow constraints when all event success probabilities are set to one. 
Thus, $R^*$ represents the best performance permitted by the task structure in the absence of execution failures.

\begin{table}[b]
  \centering
  \caption{
  Agent assignments during Chess-Maze training.
  Each entry specifies the agent (\pcKnight, \pcKing, or \pcRook) assigned to an indexed event and its estimated success probability. 
  Events not required by the current high-level plan are left blank. 
  Red values indicate an empirical success probability that is well below the algorithm-specified requirement (between 0.97 and 1 throughout).
  Shaded entries indicate that a new agent has been assigned to the event since it was last required. 
  }
  \label{tab:assignments}
  \definecolor{cmFail}{HTML}{B3261E}
\definecolor{cmSwitch}{HTML}{DCE6F2}
\setlength{\tabcolsep}{3.2pt}
\renewcommand{\arraystretch}{1.1}
\begin{tabular}{@{}c *{5}{c}@{}}
\toprule
 & \multicolumn{5}{c}{Total Elapsed Training Steps ($\times 10^6$)} \\
\cmidrule(l){2-6}
Event & 0.94 & 1.31 & 1.88 & 2.50 & 2.75 \\
\midrule
\multicolumn{6}{@{}l}{\emph{Upper path} ($u_I \xrightarrow{Y} u_1 \to \cdots$)} \\
$Y_1$ & \pcKing\,\textcolor{cmFail}{\textbf{0.01}} & \textcolor{black!30}{--} & \textcolor{black!30}{--} & \textcolor{black!30}{--} & \textcolor{black!30}{--} \\
$R_1$ & \pcKnight\,0.98 & \textcolor{black!30}{--} & \textcolor{black!30}{--} & \textcolor{black!30}{--} & \textcolor{black!30}{--} \\
$Y_3$ & \pcKing\,1.00 & \textcolor{black!30}{--} & \textcolor{black!30}{--} & \textcolor{black!30}{--} & \textcolor{black!30}{--} \\
$B_2$ & \pcKnight\,0.99 & \textcolor{black!30}{--} & \textcolor{black!30}{--} & \textcolor{black!30}{--} & \textcolor{black!30}{--} \\
$T_1$ & \pcKing\,\textcolor{cmFail}{\textbf{0.55}} & \textcolor{black!30}{--} & \textcolor{black!30}{--} & \textcolor{black!30}{--} & \textcolor{black!30}{--} \\
\midrule
\multicolumn{6}{@{}l}{\emph{Lower path} ($u_I \xrightarrow{B} u_6 \to \cdots$)} \\
$B_3$ & \textcolor{black!30}{--} & \pcKing\,\textcolor{cmFail}{\textbf{0.00}} & \cellcolor{cmSwitch}\pcRook\,\textcolor{cmFail}{\textbf{0.00}} & \cellcolor{cmSwitch}\pcKnight\,0.99 & \pcKnight\,0.99 \\
$R_2$ & \textcolor{black!30}{--} & \pcKing\,\textcolor{cmFail}{\textbf{0.00}} & \cellcolor{cmSwitch}\pcKnight\,1.00 & \pcKnight\,1.00 & \pcKnight\,1.00 \\
$B_5$ & \textcolor{black!30}{--} & \pcRook\,\textcolor{cmFail}{\textbf{0.00}} & \pcRook\,\textcolor{cmFail}{\textbf{0.00}} & \textcolor{black!30}{--} & \cellcolor{cmSwitch}\pcKnight\,0.98 \\
$Y_5$ & \textcolor{black!30}{--} & \pcKing\,0.98 & \pcKing\,1.00 & \textcolor{black!30}{--} & \pcKing\,1.00 \\
$Y_4$ & \textcolor{black!30}{--} & \textcolor{black!30}{--} & \textcolor{black!30}{--} & \pcKing\,1.00 & \textcolor{black!30}{--} \\
$B_4$ & \textcolor{black!30}{--} & \textcolor{black!30}{--} & \textcolor{black!30}{--} & \pcRook\,\textcolor{cmFail}{\textbf{0.02}} & \textcolor{black!30}{--} \\
$P_1$ & \textcolor{black!30}{--} & \pcKnight\,1.00 & \pcKnight\,1.00 & \pcKnight\,1.00 & \cellcolor{cmSwitch}\pcRook\,1.00 \\
$T_2$ & \textcolor{black!30}{--} & \pcKing\,\textcolor{cmFail}{\textbf{0.71}} & \pcKing\,0.99 & \pcKing\,0.99 & \pcKing\,0.99 \\
\bottomrule
\end{tabular}

\end{table}

\begin{figure*}[!t]
    \input{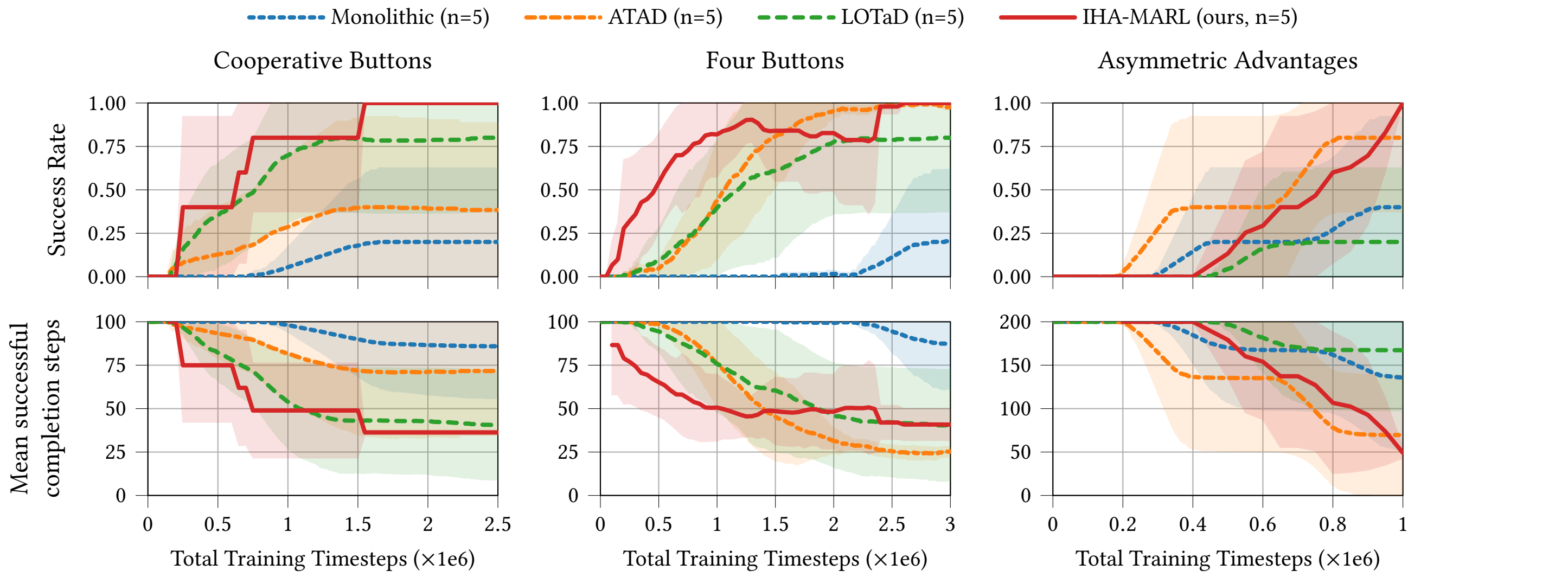}
    \caption{
    Evaluation performance of IHA-MARL (ours), LOTaD, ATAD,
and Monolithic policies on Cooperative Buttons, Four Buttons,
and Asymmetric Advantages versus cumulative training timesteps.
Top: proportion of evaluation episodes completing the team task.
Bottom: mean completion steps; failures contribute the episode horizon.
Curves show smoothed means across five independent
training seeds; shaded bands denote pointwise 90\%
Student-$t$ confidence intervals across seeds.
After training terminates, the final evaluation estimate is
held constant to the end of the comparison budget.
    }
    \label{fig:benchmark-results}
\end{figure*}

Extending the flow formulation of \citet{VCRL} to the heterogeneous multi-agent setting, we choose the assignments, contract requirements, and occupancies by solving
\begingroup
\setlength{\jot}{2pt}
\begin{align*}
\min_{z,p,d}\quad
&\sum_{x\in\Sigma}\left(p_x-\sum_{i\in I}\sigma_{i,x}z_{i,x}\right)\\
\mathrm{s.t.}\quad
&\sum_{i\in I}z_{i,x}\leq1
&&\forall x\in\Sigma,\\
&\sum_{i\in I}\sigma_{i,x}z_{i,x}\leq p_x\leq\sum_{i\in I}z_{i,x}
&&\forall x\in\Sigma,\\
&\sum_{i\in I}z_{i,x}=0\;\Longrightarrow\;d_x=0
&&\forall x\in\Sigma,\\
&p_x\leq\sigma_{i,x}+(1-\sigma_{i,x})(1-z_{i,x})
&&\forall(i,x)\in C,\\
&Az\leq b,\qquad z_{i,x}\in\{0,1\}.
\end{align*}
\begin{gather*}
\sum_{x\in\Sigma\cup\{\epsilon\}}r_xd_xp_x
\geq p_{\mathrm{team}}R^*,\\
\sum_{x\in\operatorname{out}(u)}d_x
-\gamma
\sum_{\substack{x\in\operatorname{in}(u)\\ \operatorname{src}(x)\neq u_F}}
d_xp_x
=
\mathbf{1}\{u=u_I\},
\qquad \forall u \in U,\\
p_\epsilon=1,\quad
d_x\geq0,\quad
p_x\in[0,1].
\end{gather*}
\endgroup

We solve this optimization problem using Gurobi, as in \cite{VCRL}.

Here, $\operatorname{out}(u)$ and $\operatorname{in}(u)$ denote the indexed events leaving and entering automaton state $u$, respectively, $u_I$ is the initial state, and $\gamma\in(0,1]$ is the discount factor. 
The linear constraints $Az\leq b$ encode the admissible agent assignments, while $C$ contains agent--event pairs whose policies are treated as converged.

The objective minimizes the total improvement requested beyond the current performance of the selected policies, while the reward and flow constraints ensure that the resulting atomic contracts achieve at least $p_{\mathrm{team}}R^*$. 
The optimization specifies target requirements; the trainer iteratively improves the selected policies toward these targets, with achieved performance assessed by evaluation.
For a selected converged pair $(i,x)\in C$, the upper and lower bounds force $p_x=\sigma_{i,x}$, preventing the optimizer from requesting further improvement from that policy.

After each solve, up to $k$ non-converged selected pairs $(i,x)$ with the largest positive deficits $p_x-\sigma_{i,x}$ are chosen for further training.
Each selected policy is initialized if necessary and trained for a batch of up to $N_{\mathrm{training}}$ epochs, capped by its remaining per-pair budget of $N_{\max}$ epochs.
The resulting $\sigma_{i,x}$ values are then updated and the optimization is solved again. 
Once a policy reaches $N_{\max}$ training epochs, its pair is added to $C$.
Training stops when no selected pair has a positive deficit beyond the evaluation tolerance, the requirements problem becomes infeasible, or the total training budget is exhausted.
The dispatching policy is recovered from the resulting occupancies by normalizing the positive $d_x$ values over the events enabled at each automaton state.

\section{Experimental Results}

We evaluate IHA-MARL as described above in our Chess-Maze example and compare its performance with the Monolithic, ATAD, and LOTaD baselines on Cooperative Buttons, Four Buttons, and Asymmetric Advantages. 
These cooperative multi-agent gridworlds require agents to work together to complete tasks, with some steps requiring multiple agents to act together.
The Chess-Maze experiment illustrates capability-aware assignment and training, while the benchmark comparisons measure task success and episode length. 
IHA-MARL's low-level policies use tabular Q-learning. 
Experimental settings are provided in the supplementary material.

Table~\ref{tab:assignments} reports the selected agents and their
estimated success probabilities during training in our Chess-Maze
example (Figure~\ref{fig:chessmaze}).
After approximately $0.94 \times 10^6$ nominal training steps,
the king's performance on $Y_1$ and $T_1$ remains below the contract
requirements, prompting a switch from the upper to the lower path.
With slip probability $0.2$, unintended moves can send agents into
lava, potentially contributing to the king's difficulty on $T_1$.
On the lower path, $R_2$ and $B_3$ pass to the knight, the order
of the simultaneous presses briefly reverses through $Y_4$ and
$B_4$, and assigning $P_1$ to the rook permits the knight to
perform $B_5$. The resulting team completes the maze in the
recorded rollout, illustrating capability-aware reassignment
under the coordination constraints.

Successful same-agent policies are also reused at other automaton
indices with matching event labels, followed by further training
and validation. The reassignment of task roles illustrates the
flexibility underlying component substitutability: alternative
implementations can be considered subject to the same contract
requirements and legal-assignment constraints.
Unlike the baseline implementations, IHA-MARL explicitly tracks
agent--event success estimates and uses contract requirements
to determine which policies need further training.

Figure~\ref{fig:benchmark-results} shows that IHA-MARL reaches
observed success rates of $100\%$ in all three environments at
the displayed budget endpoints, whereas baseline reliability
varies across environments. Episode lengths are comparable,
though somewhat longer in some environments, consistent with
our sequential rather than concurrent task execution.
An empirical CPU timing pilot recorded baseline training-loop times
approximately $7$--$11\times$ those of IHA-MARL; its settings and
measurements are provided in the supplementary material.

\section{Conclusions}
We present an assume/guarantee contract theory for heterogeneous multi-agent reinforcement learning (MARL), in which contracts specify subtasks and a required success probability, so that any capable agent can satisfy them.
We introduce the \textit{compositional atomic subtask automaton} (CASA) and use it to obtain a constructive decomposition of team-task automata satisfying the assumptions of Theorem~\ref{thm:big_thm} into subtasks, removing the need to search over candidate decompositions.
Across three benchmarks with five seeds each, IHA-MARL achieves observed
$100\%$ final evaluation success. In CPU timing pilots, comparable
baselines required approximately $7$--$11\times$ more training-loop time.
Future work includes allowing independent contracts to execute concurrently rather than dispatching one indexed event at a time, relaxing the assumptions that idle agents do not interfere with one another and that the team-task automaton and labeling function are given, and developing contract refinement to further support the reuse of learned components across tasks and teams.

\newpage
\bibliographystyle{ACM-Reference-Format} 
\bibliography{references}

\clearpage
\onecolumn
\appendix
\section*{Supplementary Material}
The following appendices provide the complete decomposition proof and experimental details for this paper.
\section{Complete Proof of the Automata Decomposition Theorem}
\label{sec:decomposition-proof}

For completeness, we provide the full proof of Theorem~\ref{thm:big_thm} from the
main paper.

\begin{theorem}[Automata Decomposition]
\label{thm:big_thm_supp}
Let
\[
A=(\Sigma,U,u_I,F,\to)
\]
be a finite deterministic indexed automaton with a unique sink state
$s\in F$, such that every state is reachable from $u_I$ and has a
path to $s$.
Then there exist $|\Sigma|$ automata
\[
A_i=(\Sigma_i,U_i,u_I^i,F_i,\to_i),
\]
with
\[
U_i=\{\mathrm{armed},\mathrm{active},\mathrm{resolved}\}
\qquad\text{and}\qquad
\bigcup_i \Sigma_i=\Sigma,
\]
such that
\[
A\cong \parallel_i A_i.
\]
\end{theorem}

\begin{proof}
Let
\[
\Sigma=\{\alpha_1,\ldots,\alpha_m\},
\]
where, because $A$ is indexed, each event $\alpha_i$ labels a unique
transition
\[
u_i \xrightarrow{\alpha_i} v_i.
\]

\paragraph{Construction of the local automata.}
For each indexed event $\alpha_i$, partition $U$ into
\[
U_{\mathrm{armed}}^i=s(u_i),\qquad
U_{\mathrm{active}}^i=\{u_i\},\qquad
U_{\mathrm{resolved}}^i
=
U\setminus\bigl(s(u_i)\cup\{u_i\}\bigr),
\]
where $s(u_i)$ denotes the set of states from which $u_i$ is reachable,
excluding $u_i$ itself.

Define
\[
q_i:U\to
U_i=\{\mathrm{armed},\mathrm{active},\mathrm{resolved}\}
\]
by
\[
q_i(u)=
\begin{cases}
\mathrm{armed},
    & u\in U_{\mathrm{armed}}^i,\\
\mathrm{active},
    & u\in U_{\mathrm{active}}^i,\\
\mathrm{resolved},
    & u\in U_{\mathrm{resolved}}^i.
\end{cases}
\]

We include in the alphabet of $A_i$ every indexed event whose
transition changes the $i$th partition state, together with
$\alpha_i$ itself:
\[
\Sigma_i
=
\{\alpha_j\in\Sigma :
q_i(u_j)\neq q_i(v_j)\}
\cup
\{\alpha_i\}.
\]
In particular, $\alpha_i\in\Sigma_i$ for every $i$, and hence
\[
\bigcup_{i=1}^m\Sigma_i=\Sigma.
\]

Now define
\[
A_i=(\Sigma_i,U_i,u_I^i,F_i,\to_i),
\]
where
\[
u_I^i=q_i(u_I),
\qquad
F_i=\{q_i(u):u\in F\},
\]
and, for every $\alpha_j\in\Sigma_i$,
\[
q_i(u_j)
\xrightarrow{\alpha_j}_i
q_i(v_j).
\]
Because each indexed event labels a unique transition in $A$, each
$\alpha_j\in\Sigma_i$ induces a unique transition in $A_i$.
Therefore each $A_i$ is deterministic.

Let
\[
M=\parallel_{i=1}^m A_i
\]
and define
\[
f:U\to U_1\times\cdots\times U_m
\]
by
\[
f(u)=\bigl(q_1(u),\ldots,q_m(u)\bigr).
\]
We show that $f$ defines an isomorphism between $A$ and the reachable
portion of $M$.
Since $u_I^i=q_i(u_I)$ for every $i$, we have $f(u_I)=(u_I^1,\ldots,u_I^m)$, the initial state of $M$.

\paragraph{Injectivity.}
Suppose $f(x)=f(y)$.

First suppose that $x$ is not the sink state. Since every state has a
path to the sink, $x$ is the source of some transition
\[
x\xrightarrow{\alpha_k}x'.
\]
Hence $x=u_k$, and therefore
\[
q_k(x)=\mathrm{active}.
\]
Since $f(x)=f(y)$, their $k$th coordinates agree:
\[
q_k(y)=q_k(x)=\mathrm{active}.
\]
But
\[
U_{\mathrm{active}}^k=\{u_k\},
\]
so $q_k(y)=\mathrm{active}$ implies
\[
y=u_k=x.
\]

It remains to consider the case in which $x$ is the sink.
If $y$ were not the sink, then applying the preceding argument to
$y$ would give some $k$ such that
\[
q_k(y)=\mathrm{active}.
\]
Since the active block of the $k$th partition is the singleton
$\{u_k\}$ and the sink is not $u_k$, we would have
\[
q_k(x)\neq\mathrm{active},
\]
contradicting $f(x)=f(y)$.
Thus $y=x$, and $f$ is injective.

\paragraph{Preservation of transitions.}
We next show that transitions of $A$ induce transitions of $M$.
Suppose
\[
x\xrightarrow{\alpha_j}y
\]
in $A$.

For each local automaton $A_i$, there are two cases.
If $\alpha_j\in\Sigma_i$, then by construction
\[
q_i(x)
\xrightarrow{\alpha_j}_i
q_i(y).
\]
If $\alpha_j\notin\Sigma_i$, then by the definition of $\Sigma_i$,
\[
q_i(x)=q_i(y).
\]
Therefore, by the definition of parallel composition,
\[
f(x)
\xrightarrow{\alpha_j}
f(y)
\]
in $M$.

\paragraph{No additional reachable transitions.}
Conversely, suppose
\[
f(x)\xrightarrow{\alpha_j}z
\]
in $M$.
Since $\alpha_j\in\Sigma_j$, the $j$th component must execute its
$\alpha_j$ transition.
The source of that transition is
\[
q_j(u_j)=\mathrm{active},
\]
so
\[
q_j(x)=\mathrm{active}.
\]
Because the active block of the $j$th partition is the singleton
$\{u_j\}$, it follows that
\[
x=u_j.
\]

Let $v_j$ be the unique state satisfying
\[
u_j\xrightarrow{\alpha_j}v_j
\]
in $A$.
By the forward direction already established,
\[
f(u_j)
\xrightarrow{\alpha_j}
f(v_j)
\]
in $M$.

Each $A_i$ is deterministic, and therefore their parallel composition
$M$ is deterministic: from any product state, each component has at
most one successor under a given event, and consequently so does the
product.
Hence
\[
z=f(v_j).
\]
Thus every transition of $M$ originating from a state in the image of
$f$ corresponds uniquely to the transition with the same label in
$A$.

\paragraph{Reachable states.}
By definition, automaton isomorphism here requires a bijection between
the reachable states of the two automata.
Every state of $A$ is reachable from $u_I$ by assumption.
Since transitions are preserved in the forward direction, every
$f(u)$ with $u\in U$ is reachable from
\[
f(u_I)
=
(u_I^1,\ldots,u_I^m)
\]
in $M$.

It remains to show that $M$ has no other reachable states.
We proceed by induction on the length of a path from $f(u_I)$.

The initial product state $f(u_I)$ is clearly in the image of $f$.
Now suppose that a reachable product state is of the form $f(u)$ and
that
\[
f(u)\xrightarrow{\alpha_j}z.
\]
By the converse transition argument above,
\[
u=u_j
\qquad\text{and}\qquad
z=f(v_j).
\]
Thus every reachable successor of a state in the image of $f$ also
lies in the image of $f$.
By induction, every reachable state of $M$ belongs to $f(U)$.

Since $f$ is injective, it follows that
\[
f:U\to \operatorname{Reach}(M)
\]
is a bijection.

\paragraph{Accepting states.}
Finally, we show that $f$ preserves acceptance.

If $u\in F$, then by definition of $F_i$,
\[
q_i(u)\in F_i
\]
for every $i$.
Therefore
\[
f(u)
\in
F_1\times\cdots\times F_m.
\]

Conversely, suppose $u\notin F$.
Since the unique sink state $s$ belongs to $F$, $u$ is not the sink.
Because $u$ has a path to the sink, it is therefore the source of some
transition
\[
u\xrightarrow{\alpha_k}u'.
\]
Hence
\[
q_k(u)=\mathrm{active}.
\]
The corresponding partition block is the singleton
\[
U_{\mathrm{active}}^k=\{u\}.
\]
Since $u\notin F$, no accepting state is mapped to
$\mathrm{active}$ by $q_k$, and therefore
\[
\mathrm{active}\notin F_k.
\]
Consequently,
\[
f(u)\notin F_1\times\cdots\times F_m.
\]

Thus $f$ is a bijection between the states of $A$ and the reachable
states of $M$ that preserves the initial state, labelled transitions,
and accepting states. Hence
\[
A
\cong
\parallel_{i=1}^m A_i.
\]
\end{proof}

\section{Experimental Details}
\label{sec:experimental-details}

\subsection{Environments}

\paragraph{Chess-Maze.}
The $16\times21$ map and task automaton appear in the main paper.
The king moves to adjacent cells, the knight makes chess-knight jumps,
and the rook moves along unobstructed rows or columns; all can stay.
Pressure-plate events occur on entry, while simultaneous activations
require distinct agents to hold the relevant plates. 
Release events are executed by the agent currently occupying the
corresponding plate, and only the king can reach the throne.
Walls and closed gates block movement; entering lava disables an agent.
In the reported run (seed~1), each non-stay action has probability $0.2$
of being replaced by a uniformly sampled passable adjacent move.
The scheduler table records nominal steps: training episodes multiplied
by the local horizon of 50, rather than actual transitions. The saved
team is executed for at most 1,000 ticks; the reported completion is
one demonstration rollout.

\paragraph{Cooperative Buttons and Four Buttons.}
Both use $10\times10$ grids with cardinal moves and a stay action.
Movement succeeds as intended with probability $0.98$; each perpendicular
direction has probability $0.01$. Cooperative Buttons uses three agents:
yellow and green buttons enable access for other agents, and two agents
must jointly hold the red button to allow the remaining agent to reach
the goal. Four Buttons uses two agents to activate all four buttons;
activating red before yellow irreversibly violates the task.

\paragraph{Asymmetric Advantages.}
Two cooks occupy separate sides of a $5\times9$ kitchen with a shared pot.
They add three onions, wait 20 ticks for cooking, collect soup using a
dish, and deliver it. Different distances to ingredients and delivery
locations create asymmetric roles. IHA-MARL represents cooking as an
autonomous waiting component with an analytically computed probability
of completion within its deadline, rather than a learned third agent.

\subsection{Training and Evaluation}

Each training epoch comprises one local episode.
IHA-MARL trains tabular Q-policies sequentially, with learning rate $0.2$
and constant $\epsilon$-greedy exploration $\epsilon=0.2$. Local rewards
are $1$ on completing the event and $-0.01$ otherwise. Each iteration
trains at most ten selected agent--event policies; pairs reaching their
episode limit are treated as converged. Training stops when no selected
policy needs improvement or the requirements problem is infeasible.
For learned policies, capability success after each batch is the fraction of 128 greedy
validation episodes completing the event within its local horizon,
with initial states sampled from the component's reachable entry set. These empirical estimates are
not certified worst-case probabilities. Policies for the same agent
and event label can initialize tasks at other automaton indices and are
then trained and revalidated. The target in Table~\ref{tab:settings}
is a fraction of ideal abstract discounted return, not a measured
whole-team success threshold.

The PPO baselines use the LOTaD implementation.\footnote{Learning Symbolic
Task Decompositions for Multi-Agent Teams, AAMAS 2025;
\url{https://arxiv.org/abs/2502.13376}.} Monolithic assigns each
agent the full team task; ATAD fixes one candidate decomposition; LOTaD
uses UCB to select among ten candidates. Decomposition scoring uses
size and balance weights $2$ and $0.5$, respectively; UCB uses exploration
coefficient $0.5$ and forgetting factor $1$. PPO uses shared
task-conditioned policies with two 64-unit tanh hidden layers,
rollout length 2,048, minibatches of 256,
five epochs, GAE parameter $0.95$, clipping $0.2$, value-loss coefficient
$0.5$, and gradient-norm limit $0.5$.

The comparison uses seeds 11--15. The original IHA-MARL runs have no
global training cap; plotted observations are restricted to the budgets
in Table~\ref{tab:settings}. Its counter sums physical transitions during
low-level training and excludes validation and testing. PPO counts
agent slots: three per joint step in Cooperative Buttons and two in the
other domains. Thus equal numerical budgets do not imply equal joint
interactions or wall-clock time. Chess-Maze's nominal counter is separate
from both conventions.

IHA-MARL checkpoints request 20 greedy whole-team evaluations every
50,000 training transitions, with 50 additional final tests per seed.
Baseline checkpoints use one stochastic-policy evaluation episode,
every 15,000, 10,000, and 3,000 agent slots in Cooperative Buttons,
Four Buttons, and Asymmetric Advantages, respectively. Success means
acceptance of the team task; episode lengths include failures and
retain each implementation's termination conventions.
For plotting, IHA-MARL observations are carried forward onto a 50,000-step
grid and smoothed using trailing windows of 15, 10, and 3 points,
respectively; baseline windows contain 50 observations.
Means and pointwise 90\% Student-$t$ intervals are computed across the
five smoothed seed curves. 
If training terminates due to successful trained and assigned components, the final available estimate is carried forward to the plot boundary for visualization.
Observations beyond the plotted budget are excluded.

\clearpage
\subsection{Parameters and Timing}

\begin{table}[H]
\centering
\small
\setlength{\tabcolsep}{5pt}
\caption{Parameters for the reported Chess-Maze demonstration and original
five-seed comparison. CB: Cooperative Buttons; FB: Four Buttons;
AA: Asymmetric Advantages. Horizons and capability limits are in episodes
or ticks as indicated; budgets use the counters described above.}
\label{tab:settings}
\begin{tabular}{lrrrr}
\toprule
 & Chess-Maze & CB & FB & AA \\
\midrule
\multicolumn{5}{l}{\textit{IHA-MARL}} \\
Target fraction & $0.95$ & $0.98$ & $0.98$ & $0.95$ \\
Q-learning discount & $0.95$ & $0.95$ & $0.95$ & $0.99$ \\
Team-model discount & $0.99$ & $0.95$ & $0.97$ & $0.99$ \\
Local horizon (ticks) & 50 & 100 & 100 & 200 \\
Episodes per capability batch & 1,250 & 1,000 & 1,000 & 1,000 \\
Maximum episodes per capability & 5,000 & 10,000 & 20,000 & 20,000 \\
Capabilities before re-solving & 10 & 10 & 10 & 10 \\
Configured team horizon (ticks) & 1,000 & 100 & 100 & 200 \\
Plotted comparison budget & --- & $2.5\times10^6$ & $3\times10^6$ & $10^6$ \\
\midrule
\multicolumn{5}{l}{\textit{PPO baselines}} \\
Learning rate & --- & $5\times10^{-4}$ & $5\times10^{-4}$ & $2.5\times10^{-4}$ \\
Discount & --- & $0.95$ & $0.97$ & $0.99$ \\
Entropy coefficient & --- & $0.1$ & $0.1$ & $0.01$ \\
Target KL & --- & $0.05$ & $0.1$ & None \\
\bottomrule
\end{tabular}
\end{table}

Separate timing pilots use seeds 11--13, a nominal counter limit of
100,000, and unchanged learning settings. Jobs run sequentially on an
Apple M1 CPU, with PyTorch, BLAS/OpenMP, and Gurobi limited to one thread.
Each checkpoint uses one evaluation episode, approximately every
20,000 counter units; IHA-MARL also performs one final test and retains
its 128-episode capability validations.
Table~\ref{tab:timing} reports median elapsed training-loop time including
checkpoint evaluations. PPO's timer covers \texttt{learn}; IHA-MARL's
also includes assignment optimization, capability validation,
serialization, and an execution check. Process startup is excluded.
PPO completes rollouts beyond the cap, and IHA-MARL can stop early
(as occurred for one Cooperative Buttons seed).
These pilots measure implementation cost under the stated counters,
rather than time to a specified success rate; they do not establish
full-budget speedups or isolate assignment from the Q-learning/PPO
learner difference.

\begin{table}[H]
\centering
\small
\caption{Median training-loop time in seconds over three timing-pilot seeds.}
\label{tab:timing}
\begin{tabular}{lrrrr}
\toprule
Environment & IHA-MARL & Monolithic & ATAD & LOTaD \\
\midrule
Cooperative Buttons & 1.10 & 10.29 & 10.77 & 10.20 \\
Four Buttons & 1.89 & 14.09 & 15.70 & 14.72 \\
Asymmetric Advantages & 132.02 & 1,400.38 & 1,447.00 & 1,368.63 \\
\bottomrule
\end{tabular}
\end{table}

\end{document}